# Fastidious Attention Network for Navel Orange Segmentation

SUN Xiaoye[1,2], LI Gongyan[1] and XU Shaoyun[1]

1. Institute of Microelectronics of Chinese Academy of Sciences, Beijing 100029, China
2. University of Chinese Academy of Sciences, Beijing 100049, China

sunxiaoye16@mails.ucas.ac.cn, {ligongyan,xushaoyun}@ime.ac.cn

**Abstract**— Deep learning achieves excellent performance in many domains, so we not only apply it to the navel orange semantic segmentation task to solve the two problems of distinguishing defect categories and identifying the stem end and blossom end, but also propose a fastidious attention mechanism to further improve model performance. This lightweight attention mechanism includes two learnable parameters, activations and thresholds, to capture long-range dependence. Specifically, the threshold picks out part of the spatial feature map and the activation excite this area. Based on activations and thresholds training from different types of feature maps, we design fastidious self-attention module (FSAM) and fastidious inter-attention module (FIAM). And then construct the Fastidious Attention Network (FANet), which uses U-Net as the backbone and embeds these two modules, to solve the problems with semantic segmentation for stem end, blossom end, flaw and ulcer. Compared with some state-of-the-art deep-learning-based networks under our navel orange dataset, experiments show that our network is the best performance with pixel accuracy 99.105%, mean accuracy 77.468%, mean IU 70.375% and frequency weighted IU 98.335%. And embedded modules show better discrimination of 5 categories including background, especially the IU of flaw is increased by 3.165%.

**Keywords**— Semantic Segmentation, Fastidious Attention, Navel Orange

## 1 Introduction

The navel orange external quality and consistency affects its commercial value[1] and market competitiveness[2]. While computer vision technology can not only replace visual inspection by workers to improve efficiency, but also evaluate the homogeneity of the external quality of navel orange with the advantages of nondestructive and objective[3]. And in recent years, deep learning is shining in the field of computer vision. Therefore, based on deep learning, we try to solve some difficult problems of navel orange segmentation to improve performance of external quality inspection.

The stem end and blossom end of the navel orange are easily misjudged into defects, so distinguishing them is the key to the defect segmentation of the navel orange. And it is a difficult task to further identify the types of defects based on the detected defects. Because target structures on navel oranges typically present intra and inter-class diversity on size, shape and texture. As shown in Fig. 1, we show four commercial categories of navel orange: stem end, blossom end, flaw and ulcer.

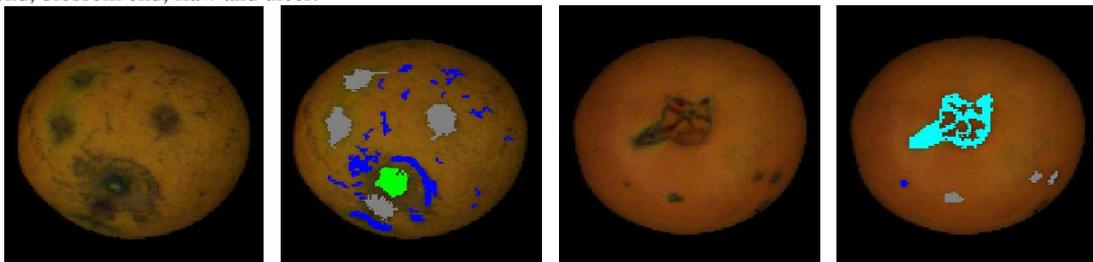

(a) The stem end of navel orange with flaw and ulcer   (b) The mask annotation of (a)   (c) The blossom end of navel orange with flaw and ulcer   (d) The mask annotation of (c)

Fig. 1 Green, light blue, dark blue, and gray masks are stem end, blossom end, flaw, and ulcer of navel orange.

There are a lot of excellent work[1]–[4] for orange to segment defects, but not to identify them. And many of them[3], [5], [6] only focused on the effect of the stem end on navel orange defects, but navel orange has obvious blossom end which cannot be ignored. First, it is important to distinguish between defect types. An individual navel orange may have various defect types. As shown in Fig. 1 (a), navel orange has both flaw and ulcer. And defects like flaw may only affect the consistency at surface of navel orange, but ulcer is likely to be zero tolerance for both consumers and producers[2]. Then we found that there are many navel orange sampling images with similar blossom end as shown in Fig. 1 (b), which are very similar to the defect.

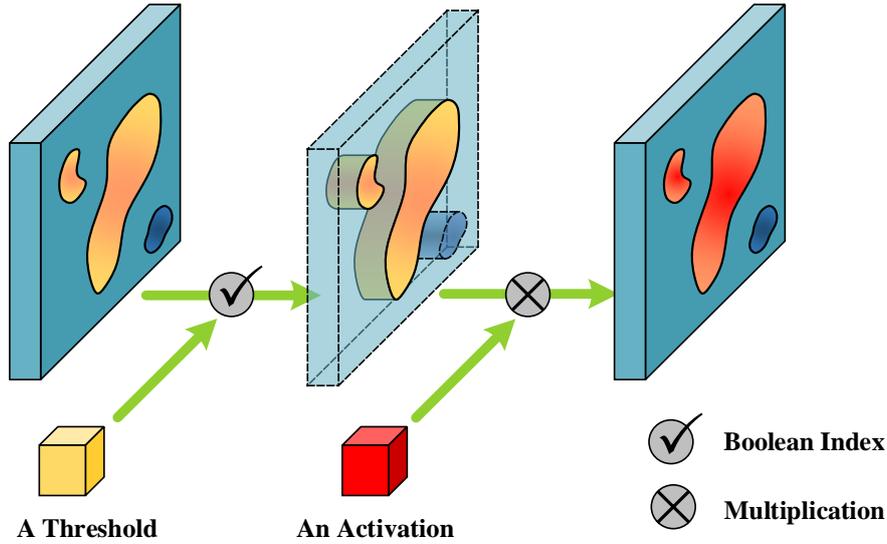

Fig. 2 Fastidious excitation: the region of focus (solid orange part) is selected from a feature map of channel c by a threshold. The area is then uniformly excited by multiplied a weight, while the rest (dotted translucent part) remain constant. Threshold and weight are learnable parameters.

So, we mainly address these two problems: identify the type of segmentation defects and distinguish the stem end and blossom end of navel orange. We further convert this problem into semantic segmentation for stem end, blossom end, flaw, and ulcer of navel orange.

Deep learning demonstrate large potential and even outperform the state-of-the-art not only in many visual tasks like object detection, object recognition, image retrieval, target tracking, but also in a variety of agricultural scenarios. A survey[7] of deep learning algorithms used in agricultural applications demonstrates beyond the performance of traditional image processing. On the one hand, hand-engineered components require considerable time, but an effort that takes place automatically in deep learning[7], [8]. On the other hand, deep learning model is robust under challenging conditions such as illumination, complex background, different resolution, size, scale, occlusion, variation and orientation of the images[8]–[10]. Specifically, before the outbreak of deep learning, the uneven lightness distribution on the surface of oranges has been an important factor affecting model performance and much work[3], [6] studies its. However, the deep learning model has better performance under varying illumination conditions, and because of its powerful feature extraction ability, it can better represent for orange images present faint defect characters or inhomogeneous surface.

Because segmentation targets of orange are high similarity with medical images like segmentation for vessels, tissues, nerves, tumors, and so on. And U-Net[11] is the representative application of CNN in the field of medical images, and many subsequent medical image segmentation architectures[12]–[15] are inspired from it. So, we not only apply U-Net to navel orange segmentation tasks, but also further propose models with better performance.

Capturing long-range dependencies between feature maps is of central importance in deep neural networks[16]. Because it aggregates long-range contextual information, so improving discriminant feature representations[17]. Attention modules focus on emphasizing important local regions captured in local features and filtering irrelevant information transferred by global features, thus it can model long-range dependencies[18]. In particular, the self-attention mechanism[19], which is originally applied to machine translation, is increasingly being applied in image vision flied[16], [17], [20]. However, this method still have high computation complexity and occupy a huge number of GPU memory[20].

Different from the previous attention mechanism, we propose a fastidious attention mechanism. Inspired by the lightweight gating mechanism of the SE block[21], which learns a parameter to excite for each channel feature map, we further make attention fastidious based on it. A single-channel feature map not only corresponds to a learnable activation, but also a learnable threshold. The threshold divides the feature map into foreground and background, and it considers that each pixel of the feature map selected as the foreground has a dependency relationship. Therefore, it is easy to select feature pixels with long-range dependence. As shown in Fig. 2, a 1×1 threshold in fastidious attention mechanism generates under broadcasting boolean masks of the same size as the feature map. Boolean masks fancy index feature pixels in the feature map, then these pixels are excited by an activation. We define such a process as fastidious excitation.

In order to obtain the learnable parameters of thresholds and activations in the fastidious excitation, we design a fastidious self-attention module (FSAM) and a fastidious inter-attention module (FIAM). In FSAM, the learnable parameters of the feature map to be excited are trained from itself. The learnable parameters in FIAM are trained from the deepest feature map in downsampling, and the parameters provided to each layer of fastidious excitation are interdependent.

We construct the Fastidious Attention Network (FANet), which uses U-Net as the backbone and embeds FSAM and FIAM. Compared with some state-of-the-art models like U-Net[11], FC-DenseNet[22], and Att U-Net[12], the experimental results show that FANet achieves the best performance. We not only adopt the common semantic segmentation metrics and IU of each category for model evaluation, but also use the precision matrix, recall matrix, and F1-score matrix to further analyze the error between the categories. Then, we performed ablation and replacement studies on FSAM and FIAM, respectively, to verify the effectiveness of the module. Finally, the learning parameters of the attention module in the entire test set are counted, and some input and output feature maps of these modules are picked out to visualize. Furthermore, it proves that the fastidious attention mechanism can model long-range dependence. The main contributions in this paper are summarized as follows:

(1) We propose a fastidious attention mechanism and design fastidious self-attention module (FSAM) and fastidious inter-attention module (FIAM) based on it to improve the performance of the U-Net model in navel orange segmentation tasks;

(2) We design fastidious attention network (FANet) based on convolutional neural network to address defects segmentation difficult problems of navel orange: identify defects, stem end and blossom end.

(3) Our model shows the best performance on our navel orange dataset compared to the state-of-the-art model.

This paper first introduces the segmentation of surface defect in navel oranges and some of the current segmentation techniques in Section 2. Then, Section 3 shows the architecture we proposed and the settings for its training. In Section 4, we present navel orange dataset and the details of the model implementation, and compare it with the state-of-the-art models. Next, in Section 5 verifies the validity of the design and improvement of the model. The final section 6 concludes this paper.

## 2 Related Work

**Segmentation on Orange** There are many orange segmentation based on methods such as edge detection, thresholding, and color based segmentation. Dael et al.[23] uses automatic threshold to segment radiographs with healthy and affected tissue, and derive feature from the segmented to classify the images using a naive Bayes or KNN classifier. Based on multispectral PC images, Li et al.[2] proposed an improved watershed segmentation method to segment decay regions in orange, and this method has the function with morphological gradient reconstruction, marker extraction as well as image amendment. Thendral et al.[24] make a comparative analysis of edge and color based segmentation for orange obtained under natural lighting conditions. To detect individual fruits and obtain pixel wise mask for each detected fruit in an image captured in orchard environments, Ganesh et al.[25] present a multi-modal deep learning approach by augmenting Mask R-CNN to include HSV input data and it has better performance under natural lighting conditions.

**Semantic Segmentation** FCN[26], which can make dense predictions for pixel-wise classification and preserve the contextual spatial information by fuse the output with shallower layers 'output, is a high-impact CNN-based segmentation models. Then, SegNet[27] uses the complete encoder-decoder mode to map the low-resolution encoder feature to full input resolution feature maps and its decoder uses pooling indices. Next, U-Net[11] combines the high-resolution feature map of the encoder with the up-sampled output to assemble a more precise output. FC-DenseNet[22] introduces Dense Block in both encoder and decoder to improve efficiency in the parameter usage and feature reuse. Chen et al. propose DeepLabv3[28] to combine both dilated convolutions and feature pyramid pooling to embed contextual information.

**Self-attention Modules** It is effective for self-attention mechanism to construct long-range dependencies by capturing contextual information. Att U-Net[12] proposes a self-attention gating module, which integrates input features and the features from coarser scale to obtain spatial attention coefficients, then the input features are scaled with this coefficients. Based on the self-attention mechanism, DANet[17] introduces a position attention module and a channel attention module to model the semantic interdependencies in spatial and channel dimensions respectively. CCNet[20] designs an efficient way to harvest the contextual information of its surrounding pixels on the criss-cross path. However, Self-attention still has some computational complexity. SE block[21] utilize lightweight gating mechanism to model channel-wise relationships in a computationally efficient manner.

Different from these work, we propose a simple attention mechanism inspired by the SE block. we get threshold from channel-wise relationships to define long-range dependencies, and design two types of modules to obtain threshold and attention coefficients.

## 3 Fastidious Attention Network

Fastidious attention module composes of fastidious excitation part and learnable parameters part. In this section we first introduce fastidious excitation. Then, we show two schemas of learnable parameters part for obtaining the thresholds and activations. Next, present the aggregation between module and U-Net. Finally, give the detailed network configurations for the training.

### 3.1 Fastidious Excitation

To capture long-range contextual information for learn better feature representation in orange segmentation, we not only focus on dependencies between channels, but also selectively emphasizes interdependent spatial feature maps based on channel attention. The proposed simple attention mechanism is an array of thresholds that determine the excited part of the spatial feature map of each channel, and another array of parameters as activations.

One channel fastidious excitation depicted in Fig. 2. And so the 2D spatial feature maps $\mathbf{x}_c \in \mathbf{X} = \{\mathbf{x}_1, \mathbf{x}_2, ..., \mathbf{x}_C\}$ of channel $c$ correspond to a threshold $g_c \in \mathbf{G} = \{g_1, g_2, ..., g_C\}$ and an activation $s_c \in \mathbf{S} = \{s_1, s_2, ..., s_C\}$, where $\mathbf{G}$ is used to pick out the area to focus on, and $\mathbf{S}$ is used to excite the area. Fastidious excitation is a mapping of input feature map $\mathbf{X} \in \mathbb{R}^{H \times W \times C}$ to output feature map $\mathbf{X}' \in \mathbb{R}^{H \times W \times C}$:

$$x'_c(i,j) = \begin{cases} s_c \cdot x_c(i,j) & \text{if } x_c(i,j) > g_c; \\ x_c(i,j) & \text{otherwise,} \end{cases} \tag{1}$$

where $x_c(i,j)$ and $x'_c(i,j)$ refer to the $i-th$ row and $j-th$ column value of the $\mathbf{x}_c$ and $\mathbf{x}'_c \in \mathbf{X}' = \{\mathbf{x}_1', \mathbf{x}_2', ..., \mathbf{x}'_C\}$, respectively.

### 3.2 Fastidious Self-Attention

Inspired by SE block [21], its squeeze global spatial information about every channel of feature maps into an array of channel weights. And in order to obtain the thresholds and activations provided to fastidious excitation, we use global average pooling to squeeze the input $\mathbf{X}$:

$$z_c = \frac{1}{H \times W} \sum_{i=1}^{H} \sum_{j=1}^{W} x_c(i,j), \tag{2}$$

where $z_c$ is the $c$-th element of $\mathbf{Z} = \{z_1, z_2, ..., z_C\}$, and $\mathbf{Z} \in \mathbb{R}^C$. Then we learn two branches of parameters fromv $\mathbf{Z}$ through fully connected layers for $\mathbf{S}$ and $\mathbf{G}$:

$$\mathbf{S} = \sigma(\mathbf{W}_2^s \delta(\mathbf{W}_1^s \mathbf{Z})), \tag{3}$$

$$\mathbf{G} = \sigma(\mathbf{W}_2^g \delta(\mathbf{W}_1^g \mathbf{Z})). \tag{4}$$

Here $\delta$ and $\sigma$ refers to a ReLU function[29] and a sigmoid function, respectively. And there are two branches of fully connected layers with bottleneck structure[30] by reduction ratio r in each branch, and their weights are $\mathbf{W}_1^s \in \mathbb{R}^{C/r \times C}$, $\mathbf{W}_2^s \in \mathbb{R}^{C \times C/r}$, $\mathbf{W}_1^g \in \mathbb{R}^{C/r \times C}$, and $\mathbf{W}_2^g \in \mathbb{R}^{C \times C/r}$.

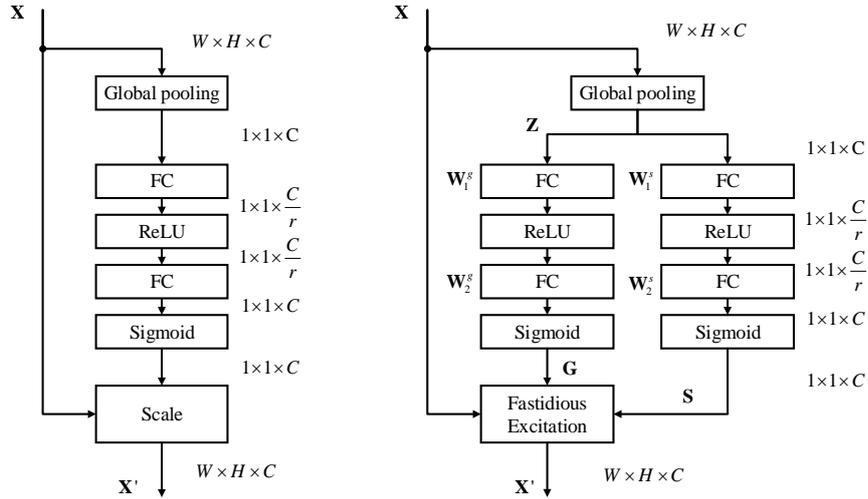

(a) The schema of squeeze and excitation module    (b) The schema of fastidious self-attention module (FSAM)

Fig. 3 Compared to SE-layer, FSAM learns $\mathbf{S}$ and $\mathbf{G}$ through the double-branch fully connected bottleneck structure for fastidious excitation.

### 3.3 Fastidious Inter-Attention

As opposed to fastidious self-attention which is based on its own feature maps $\mathbf{X}$ to obtain an array of channel weights, fastidious inter-attention provides $\mathbf{S}_n$ and $\mathbf{G}_n$ for each feature maps $\mathbf{X}_n$ based on the same feature maps $\mathbf{X}_0$, and these weights between each group are related to each other.

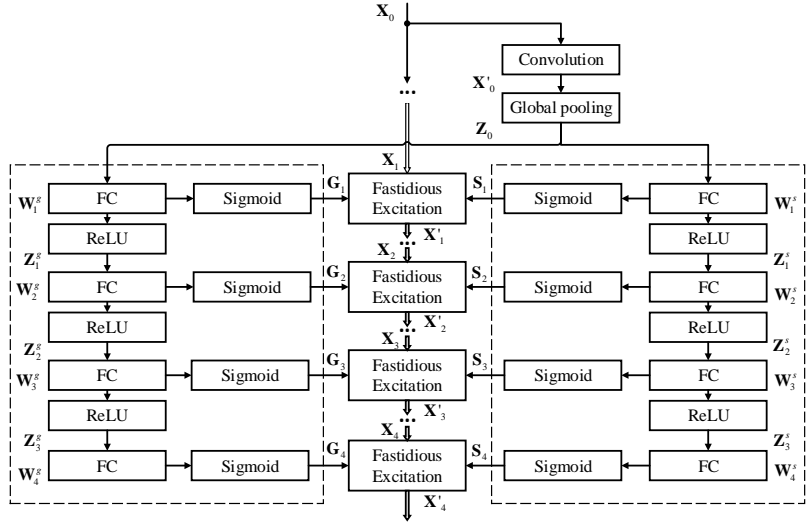

Fig. 4 The schema of fastidious inter-attention module (FIAM)

As illustrated in Fig. 4, $\mathbf{X}'_0$ after the convolution operation is applied to formula (2) to compress the global spatial information of each channel to obtain $\mathbf{Z}_0$. Then $\mathbf{Z}_0$ is the input of two fully connected networks with the same structure. And each corresponding fully connected layer outputs an array of $\mathbf{S}_n$ and $\mathbf{G}_n$ after sigmoid functions and outputs the input $\mathbf{Z}_n^s$ and $\mathbf{Z}_n^g$ of next layer after ReLU functions. This process is described as follows:

$$\mathbf{Z}_0^s = \mathbf{Z}_0^g = \mathbf{Z}_0;$$
$$\mathbf{Z}_n^s = \delta(\mathbf{W}_n^s \mathbf{Z}_{n-1}^s);$$
$$\mathbf{Z}_n^g = \delta(\mathbf{W}_n^g \mathbf{Z}_{n-1}^g); \quad (5)$$
$$\mathbf{S}_n = \sigma(\mathbf{W}_n^s \mathbf{Z}_{n-1}^s);$$
$$\mathbf{G}_n = \sigma(\mathbf{W}_n^g \mathbf{Z}_{n-1}^g).$$

### 3.4 Fastidious Attention Network

The backbone of fastidious attention net (FANet) is based on U-Net. U-Net is depicted in Fig. 5(a). For clarity, U-Net is divided into three modules Down-Conv, Up-Conv and Merge-Conv except In-Conv and Out-Conv, and an operation concatenation. Down-Conv consists of one $2\times 2$ max pooling operation with stride 2 for down-sampling and two $3\times 3$ convolutions, and each convolution followed by a batch normalization and a rectified linear unit (ReLU). And Down-Conv excluding the max pooling operation is the composition of Marge-Conv and In-Conv. Marge-Conv fuses feature map after concatenating. Then, Up-Conv consists of one interpolating operation for up-sampling and a $3\times 3$ convolution followed by a batch normalization and a ReLU. Finally, Out-Conv only consists a $1\times 1$ convolution.

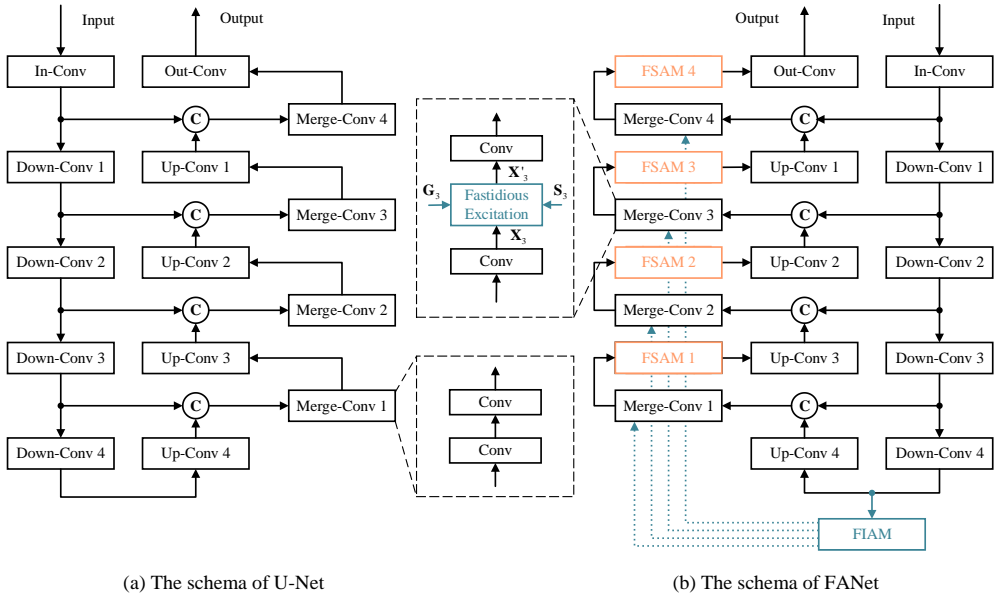

(a) The schema of U-Net  (b) The schema of FANet

Fig. 5 Architecture comparison between U-Net and FANet. The blue part is FIAM and the orange part is FSAM.

As illustrated in Fig. 5 (b), there are two types of fastidious excitation, which are behind two convolution blocks in Merge-Conv of FANet. The difference between them is that thresholds and activations are learned from different feature maps. FSAM behind each Merge-Conv learns thresholds and activations from the feature map output by Merge-Conv, and FIAM after the last Down-Conv learns thresholds and activations for fastidious excitation in all Merge-Conv.

### 3.5 Network configurations

The input resolution of FANet is $288\times288$ and there are 4 Down-Conv and Up-Conv in FANet. Therefore, there are 4 FSAMs after each Merge-Conv, and the reduction ratio r in each FSAM is 3. The number of output channels of $3\times3$ Convolution with stride 2 and padding 1 is 1.2 times the number of input channels in FIAM.

# 4 Experiment

All training and evaluating are based on our orange dataset. And performance of networks is evaluated by pixel accuracy (pixel acc.), mean accuracy (mean acc.), mean IU (mean IU) and frequency weighted IU (f.w. IU)[26]. In this section, we first introduce our orange dataset and then give the implementation details. Finally, we show the results on our dataset and present the comparisons with state-of-the-art methods.

### 4.1 Dataset

There are 6,034 color images of navel oranges collected from our orange grading machine[1]. And they are marked with the blossom end, the stem end, the flaw and the ulcer in Pascal VOC format for segmentation. Then, as shown in Table 1, these images are divided into 5,400 as training set and 634 as testing set.

Table 1 The dataset of navel orange. Here shows the number of pictures owned by each category.

|  | Train | Val | Total |
|---|---|---|---|
| Images | 5,400 | 634 | 6,034 |
| Blossom end (images) | 2,387 | 267 | 2,654 |
| Stem end (images) | 2,891 | 354 | 3,245 |
| Flaw (images) | 3,724 | 444 | 4,168 |
| Ulcer (images) | 3,991 | 494 | 4,485 |

### 4.2 Implementation Details

The implementation of FANet is based on Pytorch. During training, we set training time to 300 epochs and batch size as 4 respectively. And the network is trained from scratch using SGD with initial learning rate 0.3, 0.99 momentum and 0.0005 weight decay on Nvidia TitanX cuDNN v6.0.21 with Intel Xeon E5-2683 v3 @2.00GHz. Next, learning rate is adjusted by a cosine annealing schedule. For data augmentation, it is adopted for our dataset that the operations include random mirror (horizontally and vertically), random rotation between -180 and 180 degrees and random cropping with size $100\times100$ and padding 10. The probability of all random operations is 0.6.

### 4.3 Results on Dataset

We carry out experiments on our navel orange dataset and compare our method with existing methods. Results are shown in Table 2, which shows two-part results for segmentation metrics and each category. FANet produce dominantly performance under each metrics, especially achieving 70.375% mean IU and 77.468% mean acc.. Although all methods performed poorly in the flaw category, FANet still achieved good performance with 42.251% accuracy.

Table 2 Segmentation and per-class results on navel orange dataset. FANet outperforms existing method under each metrics.

| Method | pixel acc. | mean acc. | mean IU | f.w. IU | Background | Blossom end | Stem end | Flaw | Ulcer |
|---|---|---|---|---|---|---|---|---|---|
| PSPNet[31] | 97.681 | 40.656 | 36.872 | 95.72 | 97.677 | 25.288 | 38.625 | 5.169 | 17.602 |
| DeepLabv3 | 97.52 | 53.651 | 41.395 | 95.681 | 97.519 | 34.619 | 42.756 | 10.736 | 21.347 |
| FCN-8s | 98.477 | 66.429 | 57.294 | 97.268 | 98.506 | 52.784 | 62.099 | 22.433 | 50.648 |
| FC-DenseNet56 | 98.896 | 70.656 | 62.344 | 97.995 | 99.005 | 53.804 | 66.091 | 29.768 | 63.054 |
| FC-DenseNet103 | 98.91 | 75.159 | 65.297 | 98.052 | 99.007 | 61.417 | 67.762 | 34.203 | 64.094 |
| Att U-Net | 99.084 | 75.419 | 69.017 | 98.293 | 99.152 | 67.617 | 72.474 | 38.331 | 67.512 |
| U-Net | 99.091 | 75.889 | 69.341 | 98.305 | 99.159 | 68.189 | 72.72 | 39.086 | 67.55 |
| **FANet** | **99.105** | **77.468** | **70.375** | **98.335** | **99.171** | **69.42** | **73.146** | **42.251** | **67.884** |

---

[1] This machine is jointly developed by Jiangxi Reemoon Sorting Equipment Co., Ltd. and Institute of Microelectronics of Chinese Academy of Sciences.

We further visualize the results in Fig. 6. The first column is the RGB images input into the network, and the second column is the ground truth corresponding to each RGB image. The last three columns show the output of FANet, U-Net, and FCN8s in turn. FANet and U-Net output more details than FCN8s, and FANet is superior to U-Net in some details output.

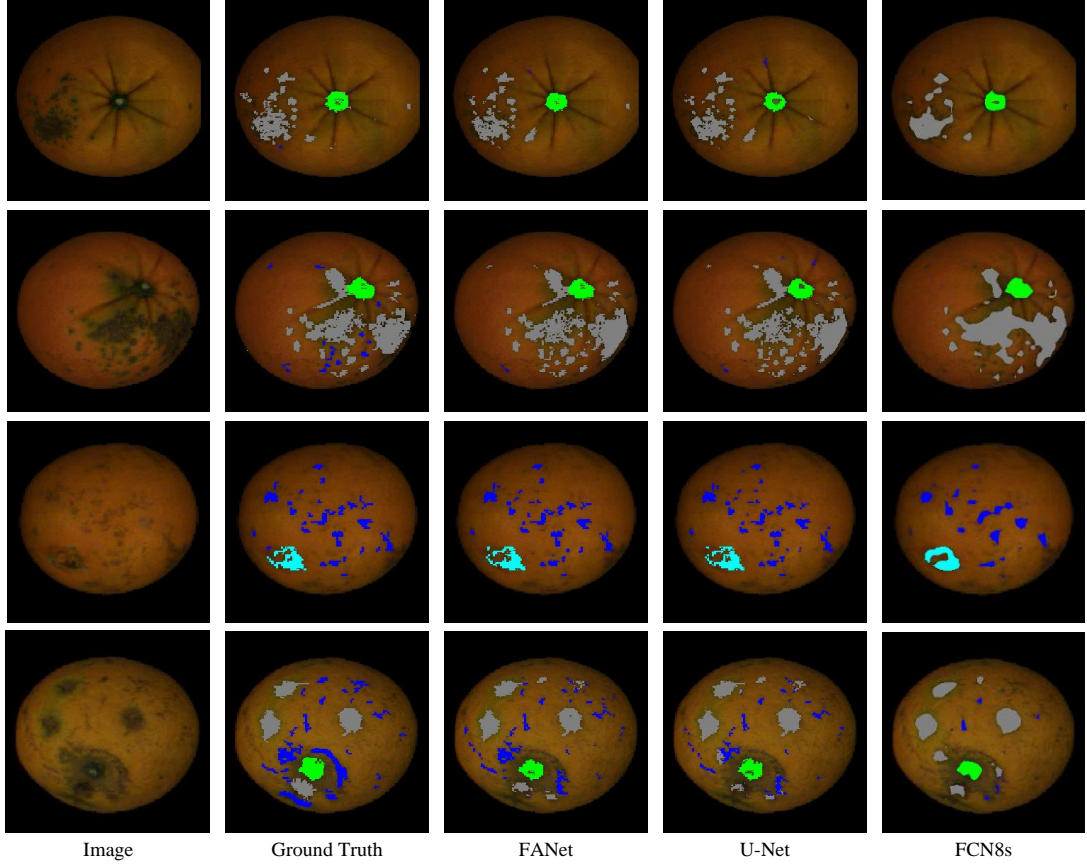

Fig. 6 Visualization results of output on navel orange dataset.

To further evaluate the performance of the network between every category, we calculate and visualize the results of precision matrix, recall matrix, and F1-score matrix in Fig. 7. The error mainly comes from the judgment between the target category and the background. And it is worth noting that the error between the two categories of flaw (num 3) and ulcer (num 4) is obvious. But FANet performs better than U-Net with flaw F1-score 59.4%.

|      | 0_p  | 1_p  | 2_p  | 3_p  | 4_p  |
|------|------|------|------|------|------|
| 0_t  | 99.4 | 10.6 | 11.3 | 19.9 | 10.2 |
| 1_t  | 0.1  | 86.2 | 0    | 0.6  | 0.2  |
| 2_t  | 0.1  | 0.3  | 87.3 | 0    | 0.2  |
| 3_t  | 0.1  | 1.8  | 0.2  | 69   | 2.9  |
| 4_t  | 0.4  | 1.1  | 1.2  | 10.5 | 86.4 |

(a) The precision matrix of U-Net

|      | 0_p  | 1_p  | 2_p  | 3_p  | 4_p  |
|------|------|------|------|------|------|
| 0_t  | 99.8 | 0    | 0    | 0    | 0.2  |
| 1_t  | 21.8 | 76.5 | 0    | 0.4  | 1.2  |
| 2_t  | 17.3 | 0.3  | 81.4 | 0    | 1    |
| 3_t  | 36.2 | 1.5  | 0.2  | 46.7 | 15.3 |
| 4_t  | 22.4 | 0.2  | 0.2  | 1.2  | 76   |

(b) The recall matrix of U-Net

|      | 0_p  | 1_p  | 2_p  | 3_p  | 4_p  |
|------|------|------|------|------|------|
| 0_t  | 99.6 | 0.1  | 0.1  | 0.1  | 0.3  |
| 1_t  | 0.1  | 81.1 | 0    | 0.5  | 0.4  |
| 2_t  | 0.1  | 0.3  | 84.3 | 0    | 0.4  |
| 3_t  | 0.2  | 1.7  | 0.2  | 55.7 | 4.9  |
| 4_t  | 0.7  | 0.3  | 0.4  | 2.2  | 80.9 |

(c) The F1-score matrix of U-Net

|      | 0_p  | 1_p  | 2_p  | 3_p  | 4_p  |
|------|------|------|------|------|------|
| 0_t  | 99.4 | 11.9 | 11.2 | 21.1 | 9.6  |
| 1_t  | 0.1  | 86.6 | 0    | 1.2  | 0.2  |
| 2_t  | 0.1  | 0.4  | 87.7 | 0    | 0.2  |
| 3_t  | 0.1  | 0.5  | 0.2  | 68.1 | 3.2  |
| 4_t  | 0.4  | 0.5  | 0.8  | 9.6  | 86.8 |

(d) The precision matrix of FANet

|      | 0_p  | 1_p  | 2_p  | 3_p  | 4_p  |
|------|------|------|------|------|------|
| 0_t  | 99.7 | 0    | 0    | 0    | 0.1  |
| 1_t  | 20.1 | 77.7 | 0    | 0.9  | 1.3  |
| 2_t  | 17.3 | 0.3  | 81.5 | 0    | 0.9  |
| 3_t  | 30.3 | 0.4  | 0.2  | 52.7 | 16.4 |
| 4_t  | 22.8 | 0.1  | 0.1  | 1.3  | 75.7 |

(e) The recall matrix of FANet

|      | 0_p  | 1_p  | 2_p  | 3_p  | 4_p  |
|------|------|------|------|------|------|
| 0_t  | 99.6 | 0.1  | 0.1  | 0.1  | 0.3  |
| 1_t  | 0.1  | 82   | 0    | 1    | 0.4  |
| 2_t  | 0.1  | 0.3  | 84.5 | 0    | 0.3  |
| 3_t  | 0.2  | 0.5  | 0.2  | 59.4 | 5.3  |
| 4_t  | 0.8  | 0.1  | 0.2  | 2.2  | 80.9 |

(f) The F1-score matrix of FANet

Fig. 7 Visualization results of precision matrix, recall matrix, and F1-score matrix. t in num_t represents ground truth, and p in num_p represents prediction. The numbers 0, 1, 2, 3, and 4 respectively represent background, blossom end, stem end, flaw and ulcer. The larger the number of diagonal elements, the better, and the smaller the value of non-diagonal elements, the better.

# 5 Discussion

In this section, we focus on the proposal attention module. First, the effects of FIAM and FSAM are studied through ablation and replacement. Then statistics and visualization for the thresholds and the activations learned by the modules are illustrated. Finally, the feature map processed in attention modules is visualized to further verify the effectiveness of the fastidious excitation.

## 5.1 Ablation and Replacement Study for Attention Modules

As reported in Table 3, For FANet without FIAM, its performance is reduced with mean IU 69.917%. And the mean IU of FANet without FSAM was reduced to 69.747%. After both FSAM and FIAM are removed, mean IU is further reduced to 69.341%. This means the network benefits from FSAM and FIAM.

Table 3 Ablation and replacement for attention modules. FANet-S represents FIAM is ablated from FANet. FANet-I represents FSAMs are ablated from FANet. FANet cuts off both FSAMs and FIAM to get U-Net. U-Net-SE is constructed by replacing FSAM in FANet-S with squeeze and excitation module(SEM).

| Network | FIAM | FSAM | pixel acc. | mean acc. | mean IU | f.w. IU | Background | Blossom end | Stem end | Flaw | Ulcer |
|---|---|---|---|---|---|---|---|---|---|---|---|
| U-Net-SE | ✗ | ✗ | 99.096 | 75.505 | 69.250 | 98.307 | 99.165 | 67.340 | 72.990 | 39.326 | 67.429 |
| U-Net | ✗ | ✗ | 99.091 | 75.889 | 69.341 | 98.305 | 99.159 | 68.189 | 72.720 | 39.086 | 67.550 |
| FANet-I | ✓ | ✗ | 99.088 | **77.639** | 69.747 | 98.316 | 99.163 | 67.684 | 72.718 | 41.481 | 67.688 |
| FANet-S | ✗ | ✓ | 99.095 | 76.662 | 69.917 | 98.311 | 99.157 | 69.076 | **73.318** | 40.444 | 67.591 |
| FANet | ✓ | ✓ | **99.105** | 77.468 | **70.375** | **98.335** | **99.171** | **69.420** | 73.146 | **42.251** | **67.884** |

To compare with squeeze and excitation module (SEM) mentioned in Fig. 3, FSAM in FANet-S is replaced by SEM. However, the mean IU of U-Net-SE is 69.25%, which is lower than FANet-S. And the U-NetE-SE with SEM is not as good as the one without it, its performance is lower than U-Net. Our network could capture long-range contextual information more effectively.

## 5.2 Learnable Parameters of Attention Module

Different images as input can get different learnable parameters (activations and thresholds), and after traversing the entire testing set, we obtain the distribution of learnable parameters for each block of FIAM and FSAM, as illustrated in Fig. 8. As Merge-Conv approaches the output (from 512 to 64 channels), the distribution of activations and thresholds is concentrated toward 0.5. As FSAM approaches output, however, the distribution of activations and thresholds is opposite to the former. In particular, the scatter of the activations of FSAM4_64 is large. We speculate because FSAM4 is next to output.

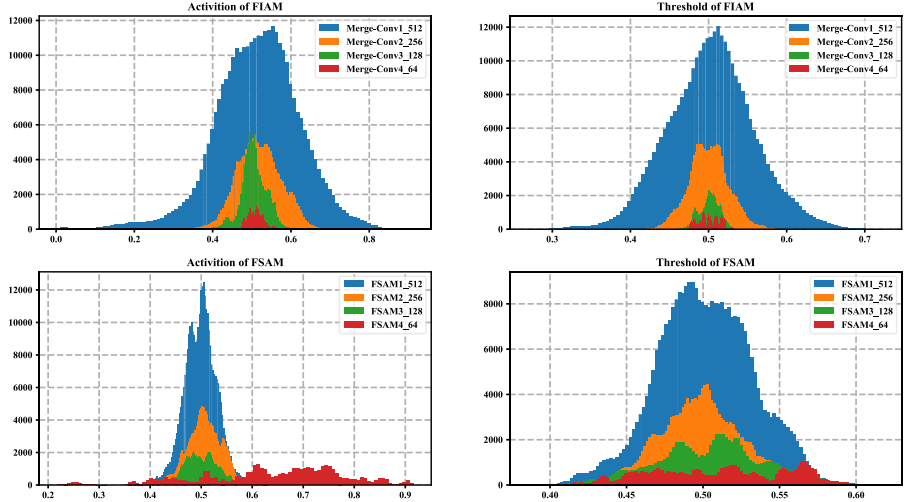

Fig. 8 The statistics of the threshold and activation of FIAM and FSAM in the navel orange testing set. For example, Merge-Conv1 in Merge-Conv1_512 corresponds to the Merge-Conv1 block in Fig. 5Fig. 5, and 512 represents the number of channels of the corresponding feature map. Its thresholds and activations are from FIAM. Similarly, FSAM2 in FSAM2_256 is the FSAM2 block in Fig. 5.

## 5.3 Visualization of Attention Module

The visualization results of FSAM and FIAM is depicted in Fig. 9. The last column demonstrates the effectiveness of fastidious excitation. The ratio contains only the value 1 and the minimum value of full image. The value 1 means that the pixel is not excited, and the minimum value is activation. The little horizontal red bar in the picture coincides with the lower end of the color bar to verify this. So, the selected white area is excited with a fixed value.

The third column shows how much the selected area is suppressed. Since both activation and threshold are output through the last sigmoid function, their range is 0 to 1. This means that the selected excited pixels are actually suppressed, and the unselected areas are the background. So, the difference between the output feature map values is smaller than the input, as illustrated in the first two columns.

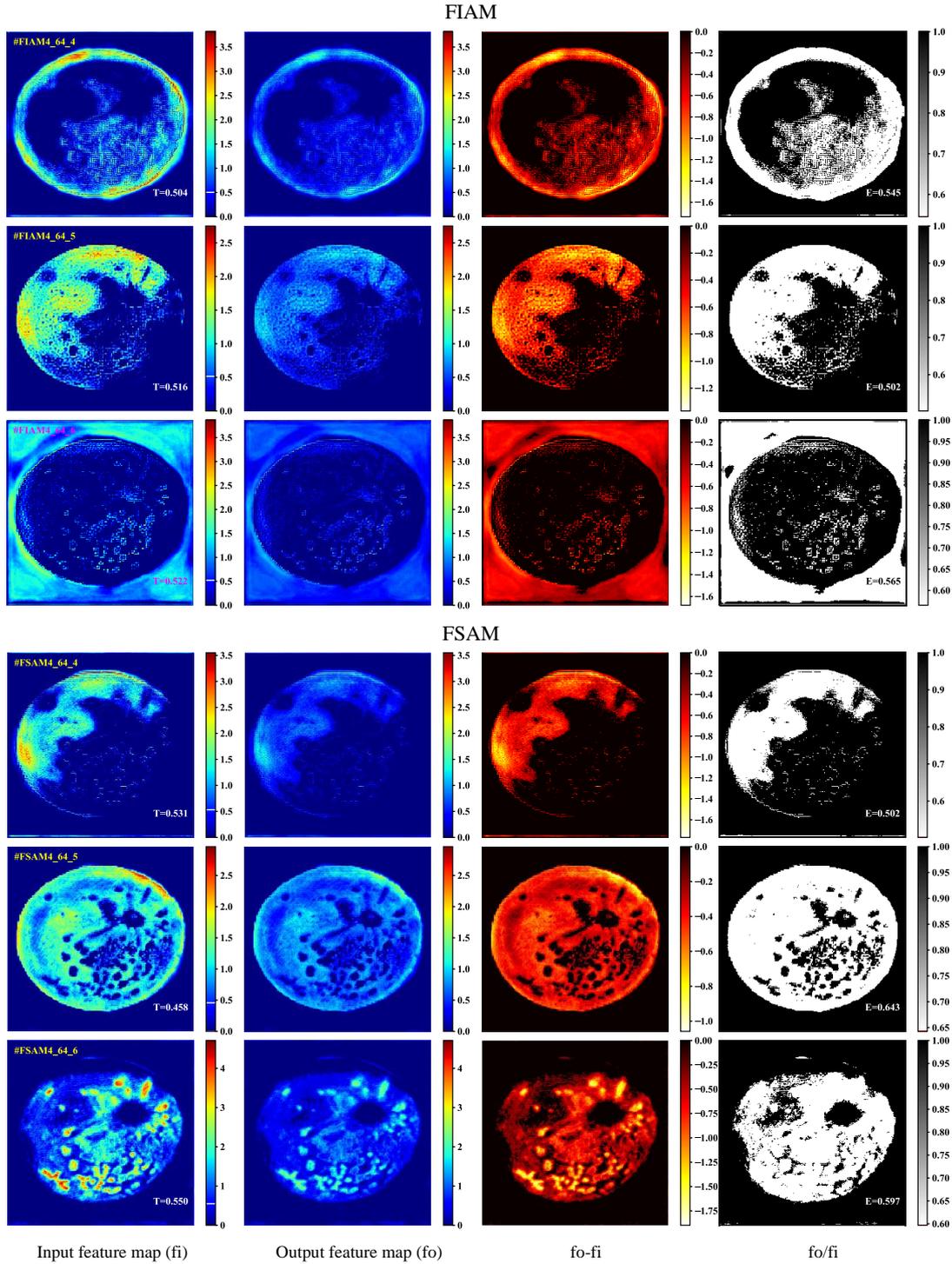

Fig. 9 The visualization results from the 4[th] to 6[th] channels of FIAM4 and FSAM4 feature maps. The first two columns are the feature map input to the attention module and the feature map output from it. The third column is the variation between the output and input feature maps. And the last column is the ratio of the output and input feature maps. T and E represent the threshold and the activation respectively. The little horizontal bars (white and red) on the color bar in the first and last columns indicate the corresponding numerical positions.

# 6 Conclusion

In this paper, we designed the Fastidious Attention Network (FANet) based on deep learning to solve two problems of navel orange segmentation: identify the type of segmented defects and distinguish the stem end and the blossom end. And we proposed a fastidious excitation mechanism, which can adaptively focus and excite local feature regions. The thresholds in this mechanism was used to select the region, and its activations was used to excite the region. Based on different feature maps used to obtain two learnable parameters of activations and thresholds, we designed two structures, fastidious self-attention module (FSAM) and fastidious inter-attention module (FIAM). Then constructed FANet, which used U-Net as the backbone and embedded these two modules. The ablation and replacement experiments showed the effects of FSAM and FIAM on model performance and their effectiveness of capturing long-range contextual information. And the visualization of feature maps further verified the effectiveness of the fastidious excitation mechanism. Compared with some state-of-the-art models like U-Net, FC-DenseNet, and Att U-Net, FANet achieved outstanding performance on navel orange dataset with pixel accuracy 99.105%, mean accuracy 77.468%, mean IU 70.375% and frequency weighted IU 98.335%. And our network showed better discrimination of 5 categories including background, especially the IU of the flaw is increased by 3.165%. We also adopted the precision matrix, recall matrix, and F1-score matrix to further analyze the error between the categories. In addition, all methods performed poorly in the flaw category, especially error between flaw and ulcer is obvious. This will be studied in future work.

# Acknowledgment

This work is supported by National Key R&D Program of China (No. 2018YFD0700300).